\documentclass[sigconf,oneside,nonacm]{acmart}
\renewcommand\footnotetextcopyrightpermission[1]{}

\AtBeginDocument{%
  }

\acmConference[ACM MM]{Make sure to enter the correct
  conference title from your rights confirmation email}{10–14 November 2026}{Rio de Janeiro, Brazil}
\acmISBN{978-1-4503-XXXX-X/2018/06}

\usepackage{graphicx}
\usepackage{booktabs}
\usepackage{adjustbox}
\usepackage{pifont}
\usepackage[table]{xcolor}
\newcommand{\cmark}{\textcolor{green!70!black}{\ding{51}}}
\newcommand{\xmark}{\textcolor{red!80!black}{\ding{55}}}
\usepackage{array}
\usepackage{multirow,tabularx,ragged2e}
\usepackage{makecell}
\usepackage{tikz}
\usepackage{xspace}
\usepackage{svg}
\usepackage{amsmath}
\usepackage{threeparttable}

\newcommand{\best}[1]{\textbf{#1}}
\newcommand{\second}[1]{\underline{#1}}

\usetikzlibrary{decorations.pathreplacing}

\begin{document}

\title{IMPACT: A Dataset for Multi-Granularity Human Procedural Action Understanding in Industrial Assembly} 



\author{Di Wen}
\email{di.wen@kit.edu}
\affiliation{%
  \institution{Karlsruhe Institute of Technology}
  \city{Karlsruhe}
  \country{Germany}
}

\author{Zeyun Zhong}
\affiliation{%
  \institution{Karlsruhe Institute of Technology}
  \city{Karlsruhe}
  \country{Germany}
}

\author{David Schneider}
\affiliation{%
  \institution{Karlsruhe Institute of Technology}
  \city{Karlsruhe}
  \country{Germany}
}

\author{Manuel Zaremski}
\affiliation{%
  \institution{Karlsruhe Institute of Technology}
  \city{Karlsruhe}
  \country{Germany}
}

\author{Linus Kunzmann}
\affiliation{%
  \institution{Karlsruhe Institute of Technology}
  \city{Karlsruhe}
  \country{Germany}
}

\author{Yitian Shi}
\affiliation{%
  \institution{Karlsruhe Institute of Technology}
  \city{Karlsruhe}
  \country{Germany}
}

\author{Ruiping Liu}
\affiliation{%
  \institution{Karlsruhe Institute of Technology}
  \city{Karlsruhe}
  \country{Germany}
}

\author{Yufan Chen}
\affiliation{%
  \institution{Karlsruhe Institute of Technology}
  \city{Karlsruhe}
  \country{Germany}
}

\author{Junwei Zheng}
\affiliation{%
  \institution{Karlsruhe Institute of Technology}
  \city{Karlsruhe}
  \country{Germany}
}
\affiliation{%
  \institution{ETH Zurich}
  \city{Zurich}
  \country{Switzerland}
}

\author{Jiahang Li}
\affiliation{%
  \institution{Karlsruhe Institute of Technology}
  \city{Karlsruhe}
  \country{Germany}
}

\author{Jonas Hemmerich}
\affiliation{%
  \institution{Karlsruhe Institute of Technology}
  \city{Karlsruhe}
  \country{Germany}
}

\author{Qiyi Tong}
\affiliation{%
  \institution{Italian Institute of Technology}
  \city{Genova}
  \country{Italy}
}

\author{Patric Grauberger}
\affiliation{%
  \institution{Karlsruhe Institute of Technology}
  \city{Karlsruhe}
  \country{Germany}
}

\author{Arash Ajoudani}
\affiliation{%
  \institution{Italian Institute of Technology}
  \city{Genova}
  \country{Italy}
}

\author{Danda Pani Paudel}
\affiliation{
  \institution{INSAIT, Sofia University}
  \city{Sofia}
  \country{Bulgaria}
}

\author{Sven Matthiesen}
\affiliation{%
  \institution{Karlsruhe Institute of Technology}
  \city{Karlsruhe}
  \country{Germany}
}

\author{Barbara Deml}
\affiliation{%
  \institution{Karlsruhe Institute of Technology}
  \city{Karlsruhe}
  \country{Germany}
}

\author{Jürgen Beyerer}
\affiliation{%
  \institution{Karlsruhe Institute of Technology}
  \city{Karlsruhe}
  \country{Germany}
}

\author{Luc Van Gool}
\affiliation{
  \institution{INSAIT, Sofia University}
  \city{Sofia}
  \country{Bulgaria}
}

\author{Rainer Stiefelhagen}
\affiliation{%
  \institution{Karlsruhe Institute of Technology}
  \city{Karlsruhe}
  \country{Germany}
}

\author{Kunyu Peng}
\authornote{Corresponding author: \texttt{kunyu.peng@kit.edu}}
\affiliation{%
  \institution{Karlsruhe Institute of Technology}
  \city{Karlsruhe}
  \country{Germany}
}
\affiliation{%
  \institution{INSAIT, Sofia University}
  \city{Sofia}
  \country{Bulgaria}
}

\renewcommand{\shortauthors}{Wen et al.}


\begin{abstract}
We introduce IMPACT, a synchronized five-view RGB-D dataset 
for deployment-oriented industrial procedural understanding, 
built around real assembly and disassembly of a commercial 
angle grinder with professional-grade tools. To our 
knowledge, IMPACT is the first real industrial assembly benchmark that jointly provides synchronized ego--exo RGB-D capture, decoupled 
bimanual annotation, compliance-aware state tracking, and 
explicit anomaly--recovery supervision within a single real 
industrial workflow. It comprises 112 trials from 13 
participants totaling 39.5 hours, with multi-route execution 
governed by a partial-order prerequisite graph, a six-category 
anomaly taxonomy, and operator cognitive load measured via 
NASA-TLX. The annotation hierarchy links hand-specific 
atomic actions to coarse procedural steps, component 
assembly states, and per-hand compliance phases, with 
synchronized null spans across views to decouple perceptual 
limitations from algorithmic failure. Systematic baselines reveal fundamental limitations that remain invisible to single-task benchmarks, particularly under realistic deployment conditions that involve incomplete observations, flexible execution paths, and corrective behavior. The full dataset, annotations, 
and evaluation code are available at 
\url{https://github.com/Kratos-Wen/IMPACT}.

\end{abstract}



\maketitle

\section{Introduction}
\label{submission:introduction}

\begin{figure*}[htbp]
    \centering
    \includegraphics[width=\textwidth, trim=0mm 100mm 200mm 0mm, clip]{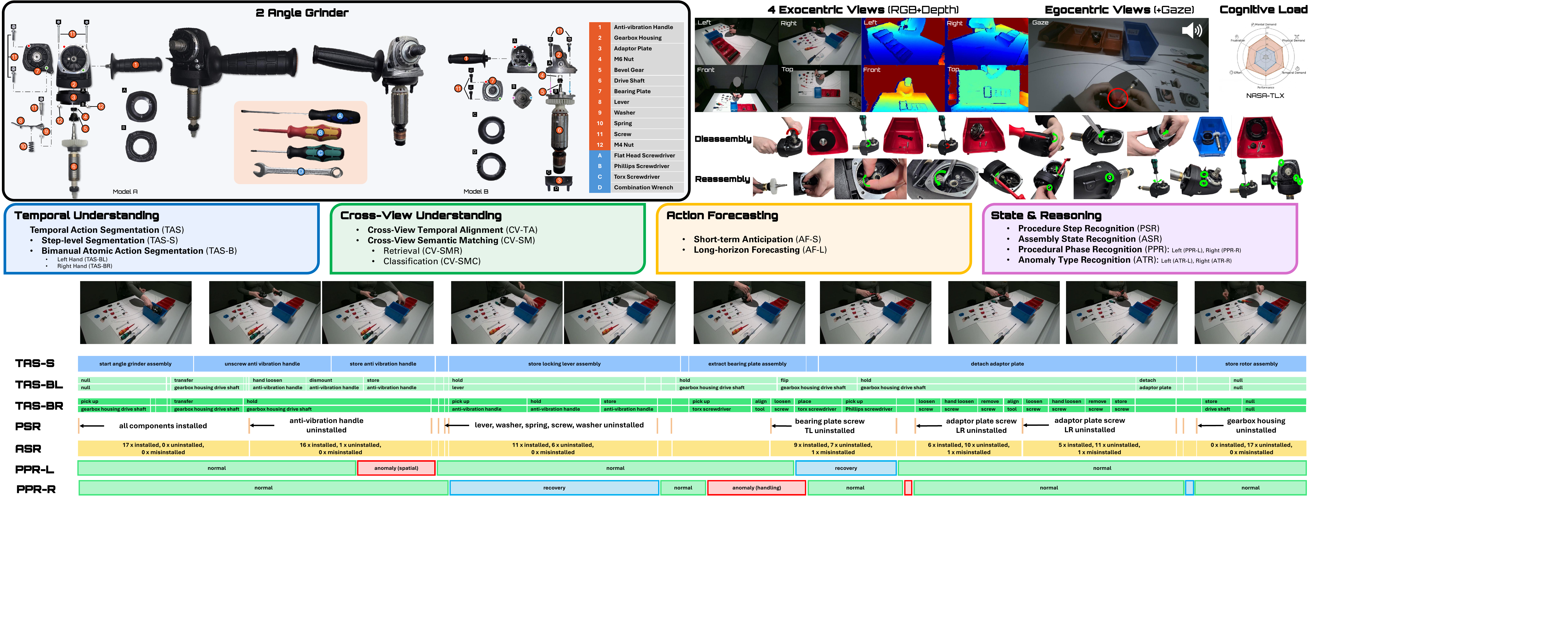}
    \vskip-3ex
    \caption{
    Overview of the IMPACT dataset and benchmark. 
    Top: recording setup with two angle grinder models, instruction manual, synchronized ego--exo RGB-D views, gaze, audio, and cognitive load. 
    Middle: task taxonomy covering temporal understanding, cross-view understanding, action forecasting, and state \& reasoning. 
    Bottom: multi-granularity annotations including step segments, bimanual actions, procedural steps, assembly states, and phases (normal, anomaly, recovery).
    }
    \Description{A multi-part figure showing an industrial assembly dataset and benchmark. The top section displays components of two angle grinders and multiple synchronized camera views, including one egocentric and four exocentric RGB-D perspectives, along with gaze visualization and a cognitive load chart. The middle section groups tasks into temporal understanding, cross-view understanding, action forecasting, and state and reasoning. The bottom section presents timeline-based annotations, including step-level segments, bimanual left and right hand actions, procedural steps, assembly states, and phase labels indicating normal, anomaly, and recovery periods.}
    \label{fig:teaser}
\end{figure*}

Intelligent industrial assistants are becoming a deployment 
reality~\cite{wen2025snap, wen2025mica}, yet the datasets 
available to train and evaluate them still under-specify the 
conditions such systems must handle in practice.
Real assembly demands more than recognizing which motion 
occurs: a system must verify whether engineering constraints 
are satisfied, track how product state evolves under occlusion 
from hands and professional tools, tolerate non-linear 
execution across valid procedural paths, and recognize and 
recover from anomalies without external intervention.
These requirements co-occur in every real deployment cycle 
and cannot be addressed independently.

Large-scale egocentric datasets~\cite{damen2022rescaling, 
grauman2022ego4d} provide semantic diversity but weak 
procedural structure. Assembly-specific 
datasets~\cite{sener2022assembly101, schoonbeek2024industreal, 
zheng2023ha} move closer, yet predominantly use proxy 
artifacts, assume near-linear execution, and evaluate 
segmentation, anticipation, or error detection in isolation.
Multi-view and ego--exo benchmarks~\cite{grauman2024ego, 
xue2023learning, park2025bootstrap} target semantic 
correspondence but not procedural compliance.
None combines view-specific unobservability, product-state 
evolution, and anomaly--recovery in a real tool-based 
industrial workflow, which is the conjunction most relevant to 
deployment.

We introduce \textbf{IMPACT}, a synchronized five-view RGB-D 
dataset built around real-world assembly and disassembly of 
a commercial angle grinder with professional-grade tools 
(Fig.~\ref{fig:teaser}).
IMPACT is organized around a single principle: the same 
recordings simultaneously support four families of 
deployment-oriented tasks --- temporal understanding, 
cross-view understanding, action forecasting, and state \& reasoning --- all under shared conditions of 
occlusion, partial-order execution, and anomaly--recovery.
Systematic baselines reveal that current methods are 
well-calibrated to stable execution yet break down 
at transitions, anomalies, and recovery, exposing 
open challenges that only become visible when these 
conditions are jointly imposed.

\noindent\textbf{IMPACT is a dataset for deployment-oriented 
industrial procedural understanding under viewpoint shift, 
partial observability, and anomaly--recovery. It makes four 
contributions:}
\begin{itemize}
\item A synchronized five-view RGB-D dataset of real 
power-tool assembly from 13 participants, with native 
multi-route execution, anomaly--recovery cycles, and 
operator cognitive metadata unavailable in any prior 
assembly benchmark.

\item A multi-granularity annotation hierarchy that links 
hand-specific atomic actions, coarse procedural steps, 
component assembly states, and compliance phases, with 
synchronized null spans across views to decouple perceptual 
limitations from algorithmic failure.

\item A unified benchmark suite where temporal 
understanding, cross-view understanding, forecasting, and 
state \& reasoning tasks all share the same recordings, 
making cross-task consistency under industrial conditions 
measurable.

\item Systematic baselines exposing three open challenges 
invisible to single-task benchmarks: the egocentric 
observability gap, the graph-structural forecasting 
ceiling, and the knowledge--execution gap in 
vision-language models.
\end{itemize}

\noindent IMPACT reframes industrial assembly understanding 
from isolated task evaluation toward structured 
reasoning over the full conjunction of conditions 
that real deployment imposes.
\section{Related Work}
\label{sec:related}

\noindent\textbf{Assembly and Procedural Benchmarks.}
Assembly benchmarks have evolved from furniture and toy 
settings~\cite{ben2021ikea, ragusa2023meccano} toward 
more realistic industrial 
scenarios~\cite{cicirelli2022ha4m, aganian2023attach, 
zheng2023ha, schoonbeek2024industreal, wang2023holoassist, 
ragusa2024enigma}. Assembly101~\cite{sener2022assembly101} 
substantially expands scale and includes mistakes and 
corrections, ATTACH~\cite{aganian2023attach} highlights 
hand-specific annotation for bimanual assembly, and 
IndustReal~\cite{schoonbeek2024industreal} introduces 
step-centric supervision with execution errors. Despite 
this progress, existing datasets either rely on simplified 
artifacts, lack explicit state-level supervision, or treat 
anomaly and recovery as implicit byproducts rather than 
first-class annotation targets. IMPACT addresses this by 
combining synchronized ego--exo RGB-D capture with 
multi-route execution, explicit anomaly--recovery 
annotation, and linked action--state supervision within 
a real commercial power-tool workflow.

\noindent\textbf{Cross-view and Ego--Exo Understanding.}
Synchronized ego--exo benchmarks~\cite{grauman2024ego, 
li2021ego, quattrocchi2024synchronization} and view-invariant 
representation learning from unpaired 
videos~\cite{xue2023learning, park2025bootstrap} have 
advanced cross-view correspondence, while 
EgoExoLearn~\cite{huang2024egoexolearn} extends this to 
procedural understanding under asynchronous viewpoints. 
However, these works primarily target semantic 
correspondence or frame-level alignment, without modeling 
procedural compliance, state evolution, or view-dependent 
observability under real occlusion. IMPACT extends this 
line by jointly evaluating instance-level alignment and 
semantic matching under synchronized multi-view execution, 
where hand and tool occlusion structurally limits what 
each viewpoint can observe.

\noindent\textbf{Forecasting, Errors, and Procedural Reasoning.}
Egocentric benchmarks establish action 
anticipation~\cite{damen2018scaling, grauman2022ego4d}, 
while procedural datasets introduce non-ideal execution 
through mistakes and 
corrections~\cite{sener2022assembly101, 
schoonbeek2024industreal, flaborea2024prego, 
lee2024error, peddi2024captaincook4d}. Recent work 
explores error detection with task graphs and multimodal 
cues~\cite{lee2025error, huang2025modeling} and object-centric 
state changes as a complementary 
signal~\cite{xue2024learning}. Forecasting, state 
tracking, and error-aware reasoning are nonetheless still 
evaluated as separate problems. IMPACT unifies them within 
a single benchmark, making structured reasoning over 
actions, state evolution, and corrective behavior jointly 
measurable for the first time.

\begin{table*}[t]
\centering
\caption{
Comparison of multiview assembly and industrial procedural datasets.
Dur$_u$ = unique duration (per execution);
Dur$_f$ = total footage (all views).
``E'' and ``X'' denote egocentric and exocentric views, respectively.
}
\vskip-3ex
\label{tab:dataset_comparison}
\small
\begin{adjustbox}{width=\textwidth}
\begin{tabular}{lcccccccccccc}
\toprule
Dataset &
Scenario &
Views &
Sensing &
\#Subj &
\#Seq &
Dur$_{u}$ &
Dur$_{f}$ &
Multi-route &
Bimanual &
Anomaly &
Human Meta &
Real Tools \\
\midrule
MECCANO~\cite{ragusa2023meccano}
& Toy model
& E1
& RGB+D+Gaze
& 20
& 20
& 6.9
& 6.9
& --
& \xmark
& \xmark
& \xmark
& \cmark \\

IKEA ASM~\cite{ben2021ikea}
& Furniture
& X3
& RGB+D+Skel
& 48
& 371
& 11.7
& 35.0
& --
& \xmark
& \xmark
& \xmark
& -- \\

Assembly101~\cite{sener2022assembly101}
& Toy vehicles
& E4+X8
& RGB
& 53
& 362
& 42.8
& 513.0
& \cmark
& \xmark
& \cmark
& \cmark\,(skill)
& \xmark \\

HA4M~\cite{cicirelli2022ha4m}
& Manufacturing
& X1
& RGB+D+IR
& 41
& 217
& 5.9
& 5.9
& \cmark
& \xmark
& \xmark
& \xmark
& \xmark \\

ATTACH~\cite{aganian2023attach}
& Cabinet
& X3
& RGB+D+IR+Skel
& 42
& 126
& 17.2
& 51.6
& \cmark
& \cmark
& \xmark
& \xmark
& \cmark \\

HA-ViD~\cite{zheng2023ha}
& Industrial assembly
& X3
& RGB+D
& 30
& 1074
& 29.0
& 87.0
& \cmark
& \xmark
& \cmark
& \cmark\,(progress/collab)
& \cmark \\

IndustReal~\cite{schoonbeek2024industreal}
& Construction toy
& E1
& RGB+D+Gaze
& 27
& 84
& 5.8
& 5.8
& \cmark
& \xmark
& \cmark
& \xmark
& \xmark \\

IKEA Ego 3D~\cite{ben2024ikea}
& Furniture
& E1
& RGB+D
& 2
& 174
& --
& --
& --
& \xmark
& \xmark
& \xmark
& -- \\

IndEgo~\cite{chavan2025indego}
& Industrial procedures
& E+X
& RGB+Gaze+Audio+Motion
& 20
& 3460E/1092X
& --
& 197E/97X
& --
& \xmark
& \cmark
& \cmark
& \cmark \\
\midrule
\rowcolor[gray]{0.92}
\textbf{IMPACT (Ours)}
& \textbf{Industrial power tool}
& \textbf{E1+X4}
& \textbf{RGB + D + Gaze + Audio}
& \textbf{13}
& \textbf{112}
& \textbf{8.0}
& \textbf{39.5}
& \cmark
& \cmark
& \textbf{$+$ recovery}
& \textbf{skill \& cognitive}
& \cmark \\
\bottomrule
\end{tabular}
\end{adjustbox}
\end{table*}

\section{The IMPACT Dataset}
\label{sec:dataset}

\begin{figure*}[t]
    \centering
    \includegraphics[width=\textwidth, trim=0mm 270mm 0mm 0mm, clip]{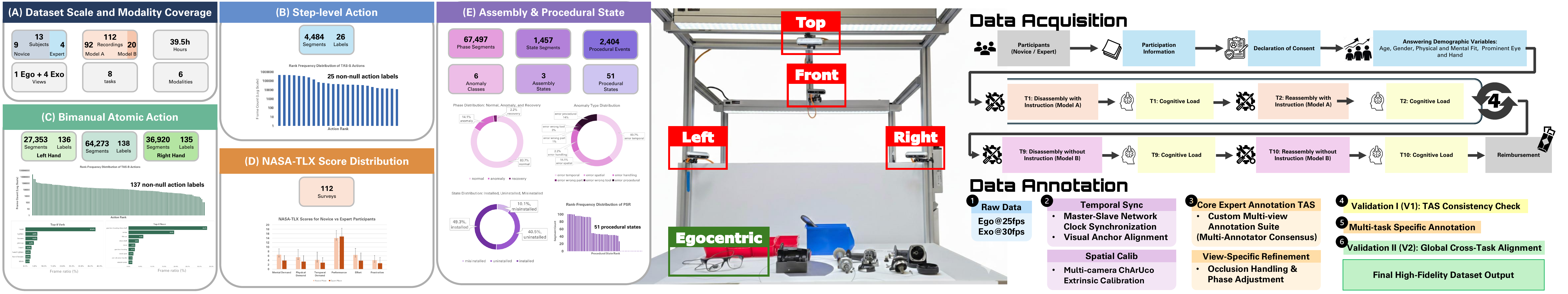}
    \vskip-3ex
    \caption{
    Data acquisition and annotation pipeline of IMPACT. Left: dataset statistics and annotation coverage. Center: synchronized multi-view setup with four exocentric RGB-D cameras and one egocentric view with gaze and audio. Right: acquisition protocol and annotation workflow, including synchronization, calibration, multi-view annotation, and multi-stage validation.
    }
    \Description{
    A figure illustrating the data collection and annotation process. 
    On the left, a workspace is shown with four external cameras labeled top, front, left, and right, and an egocentric view positioned on the table. 
    On the right, a flow diagram presents the data acquisition procedure with participants performing multiple rounds of assembly and disassembly tasks, along with cognitive load measurements. 
    The bottom part shows the annotation pipeline, including synchronization, calibration, expert annotation, validation, and final dataset generation.
    }
    \label{fig:data_pipeline}
\end{figure*}

\subsection{Assembly Task and Recording Setup}
\label{subsec:setup}

The assembly object is a commercial angle grinder requiring 
tool use, fine-grained hand coordination, and state-dependent 
procedural execution across 12 components and 4 tool types 
(Fig.~\ref{fig:teaser}). Unlike toy assembly with 
near-deterministic order, the task admits multiple valid 
execution paths governed by a partial-order prerequisite 
graph rather than a fixed sequence, directly reflecting 
real-world SOP flexibility.

The visual setup consists of four Intel RealSense D455 
exocentric RGB-D cameras (top, front, left, right) recording 
at 1280$\times$800 RGB and 1280$\times$720 depth at 30\,fps 
with overlapping fields of view, and one Tobii Pro Glasses 3 
egocentric stream at 1920$\times$1080 and 25\,fps. All 
recordings were conducted in a controlled-illumination 
laboratory on a fixed workstation. Temporal alignment uses 
master-slave network clock synchronization refined by visual 
anchor alignment; spatial calibration uses a ChArUco board 
for multi-camera extrinsic estimation. Egocentric view is 
additionally paired with gaze and audio, and 
each trial with a NASA-TLX cognitive load survey, covering 
the perceptual, procedural, and cognitive dimensions of 
real assembly in a single recording.

The dataset comprises 112 trials from 13 participants 
(7 female, 6 male; ages 19--27; 4 experts pre-trained on 
the assembly procedure, 9 novices), totaling 39.5 hours 
across two angle grinder models: 92 Model-A recordings use 
a Fein CG15-125BL and 20 Model-B recordings use a Fein 
WSG7-115A, which differs in component connectivity and 
required tool operations, providing a natural 
cross-configuration generalization axis for the S3 
evaluation split. Each participant completed repeated 
disassembly and reassembly sessions under both 
instruction-guided and free-execution conditions, yielding 
substantial variation in execution style, temporal ordering, 
and error patterns within a shared task structure.

\subsection{Multi-Granularity Annotation}
\label{subsec:annotation}

Annotation was conducted by 5 expert annotators over 7 
months, with action labels co-validated by industrial engineering and ergonomics experts to ensure fidelity, and verified by 11 
independent validators in four stages (Fig.~\ref{fig:data_pipeline}): core expert annotation in a custom multi-view interface, view-specific refinement for occlusion and phase adjustment, task-specific labeling, and two validation rounds (per-task temporal consistency; 
global cross-task alignment). Cognitive load was administered under standardized NASA-TLX protocols by certified human factors researchers.

The annotation hierarchy is designed so that each level 
addresses a distinct reasoning requirement: interaction 
dynamics, procedural structure, and physical state 
evolution.
At the \textbf{fine-grained interaction level}, hand-specific 
labels cover 137 valid action classes (22 verbs $\times$ 19 
nouns) per hand, permitting simultaneous non-null labels to 
capture coordinated bimanual behavior (27{,}353 left-hand 
and 36{,}920 right-hand segments).
At the \textbf{procedural level}, 26 step categories define 
the coarse workflow (TAS-S) and 51 completion-event 
categories form the PSR target derived from the prerequisite 
graph.
At the \textbf{state level}, the 17 component instances ternary 
annotations $\{-1,0,1\}$ (misassembled, unassembled, 
correctly assembled), yielding 1{,}457 state segments across 
51 procedural states.

Procedural phase annotations span normal (83.68\%), anomaly 
(14.09\%), and recovery (2.22\%), comprising 56{,}487 / 
9{,}512 / 1{,}498 segments respectively. Recovery is 
explicitly labeled, enabling direct evaluation of corrective behavior. Anomalies carry six non-exclusive type labels (temporal, spatial, handling, wrong part, wrong tool, procedural), reflecting that real errors commonly combine multiple failure modes. 112 NASA-TLX surveys connect observable execution to subjective workload at the trial level.

\section{Benchmark}
The annotation hierarchy described above directly instantiates four families of evaluation tasks, each targeting a distinct reasoning capability that real industrial deployment demands.
\label{sec:benchmark}
\subsection{Tasks and Evaluation Metrics}
\noindent\textbf{\textit{Temporal Understanding.}}
\textbf{Temporal Action Segmentation at step level (TAS-S)} 
assigns a coarse procedural label 
$y(t)\in\mathcal{L}_c\cup\{\varnothing\}$ per frame.
\textbf{Bimanual atomic segmentation for left and right hand 
(TAS-BL/BR)} assigns hand-specific fine-grained labels 
$z_h(t)\in\mathcal{L}_f\cup\{\varnothing\}$ for each hand 
$h\in\{L,R\}$, permitting simultaneous non-null labels to 
explicitly model synchronized bimanual interaction.
Both tasks are evaluated with frame-wise Accuracy, Edit score, 
and F1@\{10,25,50\}~\cite{lea2017temporal,farha2019ms,
sener2022assembly101}.

\noindent\textbf{\textit{Cross-View Understanding.}}
\textbf{Cross-View Temporal Alignment (CV-TA)} evaluates 
instance-level correspondence: given query segment 
$\mathbf{x}^{(u_q)}_{\tau,i}$, the model retrieves the aligned 
segment of the same occurrence from view $u_t\neq u_q$; 
same-label segments from different occurrences serve as 
negatives. We evaluate under \textbf{TA-Local} (same-trial 
candidates) and \textbf{TA-Global} (full test split), reporting 
Recall@\{1,5\} and Median Rank. The Exo$\rightarrow$Ego setting 
additionally reports Coverage (fraction of queries with 
observable egocentric targets).
\textbf{Cross-View Semantic Matching via Retrieval (CV-SMR)} 
ranks segments by semantic similarity across trials 
(Recall@\{1,5\}, mAP); \textbf{Cross-View Semantic Matching via 
Classification (CV-SMC)} predicts the verb--noun label $(v,n)$ 
(Top-1, Macro-F1), both at verb, noun, and verb--noun levels.

\noindent\textbf{\textit{Action Forecasting.}}
\textbf{Short-term Anticipation (AF-S)} predicts a future 
hand-specific interaction $(v^h,n^h)$ drawn from the TAS-BL/BR 
label space before its onset, anchored to a labeled instance 
under a fixed anticipation gap; the contralateral hand provides 
contextual input. Evaluated with mean Top-5 Recall 
(mR@5)~\cite{damen2018scaling} per hand and overall.
\textbf{Long-horizon Forecasting (AF-L)} operates on step-level 
segments from TAS-S: given $M{=}2$ observed steps, the model 
forecasts the next $Z{=}5$ steps over multiple valid execution 
paths, evaluated with ED@5 and AUED, best of $K{=}5$ 
futures~\cite{grauman2022ego4d}.

\noindent\textbf{\textit{State \& Reasoning.}}
\textbf{Procedure Step Recognition (PSR)} generalizes the 
completion-centric formulation~\cite{schoonbeek2024industreal} 
to non-unique execution orders via a prerequisite graph 
$G{=}(\mathcal{A},\mathcal{R})$; predictions are evaluated 
against the closest valid topological ordering of $G$, reporting 
Step Completion F1, Detection Delay ($\tau$), and POS.
\textbf{Assembly State Recognition (ASR)} predicts a 
component-wise state vector 
$\hat{\mathbf{s}}_t\in\{-1,0,1\}^{K}$ (misassembled, 
unassembled, correctly assembled) at each frame, evaluated with 
Macro-F1, Transition F1 (Trans-F1), and Final-State Accuracy.
\textbf{Procedural Phase Recognition for left and right hand 
(PPR-L/R)} predicts per-hand compliance phases $\{$normal, 
anomaly, recovery$\}$, where recovery denotes behavior that 
resolves a preceding anomaly; reported as Accuracy, Macro-F1, 
and class-specific F1.
\textbf{Anomaly Type Recognition for left and right hand 
(ATR-L/R)} is a multi-label diagnostic on anomalous segments, 
predicting attributes from the six-type taxonomy, decoupling compliance detection from error attribution.

\subsection{Benchmark Splits}
All splits are trial-level, co-assigning all views and 
modalities from the same execution to prevent cross-view leakage.
\textbf{S1 (IID):} label-balanced random partition.
\textbf{S2 (Cross-Subject):} held-out participants.
\textbf{S3 (Cross-Configuration):} held-out angle grinder model.
\textbf{S4 (Exo$\rightarrow$Ego):} exocentric training, 
egocentric evaluation.
Retrieval pools are restricted to the test partition; forecasting 
instances are generated post-split to prevent 
observation--target leakage.
\subsection{Baselines}
\label{sec:baselines}

Baselines are organized along the four-group taxonomy of 
Sec.~\ref{sec:benchmark}. For each group, we select models 
that span distinct architectural inductive biases, enabling 
IMPACT to diagnose which design choices matter under 
industrial procedural complexity.

\noindent\textbf{Temporal Understanding.}
For TAS-S and TAS-BL/BR, we evaluate four dense sequence 
models covering complementary design axes: long-context 
temporal modeling (LTContext~\cite{bahrami2023much}), 
query-based decoding (ASQuery~\cite{gan2024asquery}), 
diffusion-style refinement (DiffAct~\cite{liu2023diffusion}), 
and frame-action cross-attention (FACT~\cite{lu2024fact}). 
All models share the same coarse label space for TAS-S and 
the same hand-specific fine-grained label space for TAS-BL/BR.

\noindent\textbf{Cross-view Understanding.}
For CV-TA and CV-SMR/SMC, we instantiate 
I3D~\cite{carreira2017quo}, VideoMAE\,v2~\cite{wang2023videomaev2}, 
and MViTv2~\cite{li2022mvitv2} as frozen segment encoders 
with cosine-similarity $k$NN for retrieval and linear probes 
for classification. This design isolates representation 
quality from task-specific engineering, making architectural 
trade-offs directly interpretable.

\noindent\textbf{Action Forecasting.}
For AF-S, we evaluate AVT~\cite{girdhar2021anticipative} 
and ScalAnt~\cite{zhong2026scalable} as supervised baselines 
and Qwen3VL-8B~\cite{bai2025qwen3} as a zero-shot generative 
baseline, probing whether general video-language priors 
suffice for fine-grained procedural anticipation without 
task-specific training. For AF-L, we evaluate
ScalAnt~\cite{zhong2026scalable} as a visual-feature
baseline, AntGPT~\cite{qi2024antgpt} and
PALM~\cite{kim2024palm} as two-stage
recognize-then-forecast LLM pipelines, and
Qwen3VL-8B~\cite{bai2025qwen3} as a zero-shot baseline,
all operating on step-level segments from TAS-S.

\noindent\textbf{State \& Reasoning.}
For ASR, we compare MS-TCN++~\cite{li2020ms}, 
VideoMAE\,v2+Head~\cite{wang2023videomaev2}, and 
Gemini\,3.1\,Pro, spanning temporal convolution, pretrained 
encoding, and multimodal reasoning. For PSR, we evaluate 
indirect pipelines that first predict states and derive 
completion events via the prerequisite graph 
(MS-TCN++$\rightarrow$PSR; 
VideoMAE\,v2+Head$\rightarrow$PSR) against direct 
video-to-step baselines 
(STORM-PSR~\cite{schoonbeek2025learning}; 
Gemini\,3.1\,Pro), testing whether explicit intermediate 
state supervision is necessary. For PPR-L/R, we reuse 
TAS models with the compliance phase label space, 
so performance differences isolate reasoning capability 
from architectural variation.

\section{Experiments}
\label{sec:experiment}

\subsection{Training Setup and Implementation}
\label{subsec:baseline_training}

All tasks except State\,\&\,Reasoning use all five views as evaluation streams (560 streams across 112 trials); 
State\,\&\,Reasoning uses the front exocentric view only 
(112 videos) for stable workspace visibility. For representation-based baselines, backbones are frozen; features are extracted as 16-frame clips with 
stride-1 sliding windows. All models follow official implementations with default hyperparameters and 
publicly released pretrained weights; mixed-precision training 
is used where supported. VLMs are evaluated on selected benchmarks as others require per-frame dense prediction and multi-label output formats incompatible with their inference interface. All experiments run on 4$\times$A100 40\,GB GPUs. Detailed hyperparameters and training commands are provided in the supplementary. Evaluation splits follow the task structure: TAS-S, TAS-BL/BR, and CV-SMR/SMC report all four splits (S1--S4); CV-TA reports S1--S3 average for Local/Global and S4 for the Exo$\rightarrow$Ego diagnostic; AF-S, AF-L, PSR, ASR, PPR-L/R, and ATR-L/R use S1 only.


\begin{table*}[t]
\centering
\caption{Temporal Understanding Benchmarks: Step-level (TAS-S) and Bimanual Atomic Actions (TAS-B)}
\vskip-3mm
\label{tab:tas_unified}
\scriptsize
\setlength{\tabcolsep}{2.2pt}
\renewcommand{\arraystretch}{1.05}

\resizebox{\textwidth}{!}{
\begin{tabular}{l c
ccccc ccccc
ccccc ccccc
ccccc ccccc}
\toprule
\multirow{2}{*}{\textbf{Model}} &
\multirow{2}{*}{\textbf{Split}} &

\multicolumn{10}{c}{\textbf{TAS-S}} &
\multicolumn{10}{c}{\textbf{TAS-BL}} &
\multicolumn{10}{c}{\textbf{TAS-BR}} \\

\cmidrule(lr){3-12}
\cmidrule(lr){13-22}
\cmidrule(lr){23-32}

& &
\multicolumn{5}{c}{\textbf{I3D}~\cite{carreira2017quo}} &
\multicolumn{5}{c}{\textbf{VideoMAEv2}~\cite{wang2023videomaev2}} &
\multicolumn{5}{c}{\textbf{I3D}~\cite{carreira2017quo}} &
\multicolumn{5}{c}{\textbf{VideoMAEv2}~\cite{wang2023videomaev2}} &
\multicolumn{5}{c}{\textbf{I3D}~\cite{carreira2017quo}} &
\multicolumn{5}{c}{\textbf{VideoMAEv2}~\cite{wang2023videomaev2}} \\

\cmidrule(lr){3-7} \cmidrule(lr){8-12}
\cmidrule(lr){13-17} \cmidrule(lr){18-22}
\cmidrule(lr){23-27} \cmidrule(lr){28-32}

& &
\textbf{Acc$\uparrow$} & \textbf{Edit$\uparrow$} & \textbf{F1@10$\uparrow$} & \textbf{F1@25$\uparrow$} & \textbf{F1@50$\uparrow$} &
\textbf{Acc$\uparrow$} & \textbf{Edit$\uparrow$} & \textbf{F1@10$\uparrow$} & \textbf{F1@25$\uparrow$} & \textbf{F1@50$\uparrow$} &

\textbf{Acc$\uparrow$} & \textbf{Edit$\uparrow$} & \textbf{F1@10$\uparrow$} & \textbf{F1@25$\uparrow$} & \textbf{F1@50$\uparrow$} &
\textbf{Acc$\uparrow$} & \textbf{Edit$\uparrow$} & \textbf{F1@10$\uparrow$} & \textbf{F1@25$\uparrow$} & \textbf{F1@50$\uparrow$} &

\textbf{Acc$\uparrow$} & \textbf{Edit$\uparrow$} & \textbf{F1@10$\uparrow$} & \textbf{F1@25$\uparrow$} & \textbf{F1@50$\uparrow$} &
\textbf{Acc$\uparrow$} & \textbf{Edit$\uparrow$} & \textbf{F1@10$\uparrow$} & \textbf{F1@25$\uparrow$} & \textbf{F1@50$\uparrow$} \\

\midrule

\multirow{4}{*}{LTContext~\cite{bahrami2023much}}
& 1 
& \second{65.51} & \second{57.02} & \second{55.03} & \second{52.73} & \second{40.34} 
& \second{54.40} & 48.61 & 46.57 & 41.26 & 32.49 
& 48.66 & 29.48 & 27.86 & 23.04 & 13.82 
& \second{60.82} & \second{35.32} & \second{32.89} & \second{28.60} & \second{18.21} 
& 36.35 & 28.47 & 26.68 & 23.96 & 14.50 
& \second{49.23} & \second{39.64} & \second{36.61} & \second{34.01} & \second{22.71} \\

& 2 
& \best{63.75} & 55.23 & \second{53.25} & \second{50.84} & \second{38.81} 
& \second{53.52} & 50.36 & 47.86 & 43.39 & 34.09 
& \second{22.86} & 17.34 & 15.38 & 12.47 & 6.14 
& \second{23.05} & 14.09 & 11.88 & 8.51 & 4.52 
& 17.66 & 16.72 & 14.63 & 12.42 & 7.06 
& \second{20.81} & 14.73 & 12.68 & 12.64 & 7.01 \\

& 3 
& \second{35.64} & 33.55 & \second{31.58} & \second{29.40} & \second{20.54} 
& \second{27.50} & 28.78 & \second{26.96} & \second{25.29} & \second{19.42} 
& 41.10 & 14.08 & 11.98 & 9.58 & 5.16 
& \second{49.10} & 20.26 & \second{17.76} & \second{14.63} & \second{8.21} 
& \second{18.96} & 14.59 & 11.98 & 9.46 & 5.34 
& \second{27.49} & \second{16.41} & \second{13.58} & \second{12.49} & \second{6.83} \\

& 4 
& 17.83 & 15.88 & 13.73 & 12.26 & 7.45 
& 29.36 & 24.73 & 22.66 & 20.96 & 13.81 
& \second{48.12} & \second{24.61} & \second{21.55} & \second{18.16} & \second{11.72} 
& 46.37 & \second{28.61} & \second{24.86} & \second{19.98} & \second{13.65} 
& 16.56 & \second{14.65} & \second{12.37} & \second{9.88} & \second{5.86} 
& 17.79 & 15.35 & \second{12.95} & \second{10.57} & \second{6.29} \\

\midrule

\multirow{4}{*}{ASQuery~\cite{gan2024asquery}}
& 1 
& 50.39 & 56.33 & 49.76 & 44.63 & 33.89 
& 53.96 & \second{61.13} & \second{53.13} & \second{47.76} & \second{38.70} 
& \best{59.23} & \second{33.40} & \second{28.95} & \second{24.30} & \second{15.11} 
& 59.83 & 33.29 & 29.75 & 25.79 & 15.93 
& \second{45.31} & \second{32.30} & \second{29.06} & \second{24.36} & \second{15.69} 
& 47.87 & 35.35 & 31.86 & 27.12 & 18.55 \\

& 2 
& 56.37 & \best{57.61} & 51.92 & 47.02 & 36.09 
& 50.08 & \best{59.79} & \second{52.77} & \second{47.58} & \second{37.48} 
& \best{59.62} & \best{34.40} & \best{27.46} & \best{22.64} & \best{14.53} 
& \best{59.51} & \best{37.05} & \best{28.52} & \best{23.29} & \best{14.95} 
& \best{45.46} & \best{31.56} & \best{27.07} & \best{22.17} & \best{14.20} 
& \best{48.99} & \best{35.17} & \best{31.01} & \best{26.68} & \best{18.34} \\

& 3 
& \best{56.84} & \best{58.47} & \best{52.39} & \best{47.73} & \best{37.01} 
& \best{53.70} & \best{61.10} & \best{53.14} & \best{49.54} & \best{40.21} 
& \best{61.75} & \best{37.10} & \best{32.36} & \best{26.76} & \best{17.96} 
& \best{61.90} & \best{38.67} & \best{31.99} & \best{27.08} & \best{18.36} 
& \best{46.78} & \best{34.63} & \best{31.02} & \best{25.95} & \best{17.38} 
& \best{47.94} & \best{36.91} & \best{31.97} & \best{27.51} & \best{18.70} \\

& 4 
& \best{53.08} & \best{59.57} & \best{54.14} & \best{48.51} & \best{38.56} 
& \best{49.41} & \best{61.16} & \best{54.09} & \best{48.69} & \best{38.66} 
& \best{54.87} & \best{30.92} & \best{27.04} & \best{23.17} & \best{16.32} 
& \best{55.79} & \best{35.11} & \best{28.47} & \best{24.15} & \best{17.35} 
& \best{40.22} & \best{33.49} & \best{31.91} & \best{27.67} & \best{19.62} 
& \best{44.70} & \best{36.83} & \best{33.04} & \best{29.10} & \best{21.61} \\

\midrule

\multirow{4}{*}{DiffAct~\cite{liu2023diffusion}}
& 1 
& 52.51 & 48.60 & 44.26 & 38.01 & 25.28 
& 52.69 & 48.30 & 46.15 & 41.25 & 27.32 
& 18.73 & 22.34 & 17.93 & 12.37 & 5.27 
& 19.96 & 24.48 & 18.87 & 13.75 & 5.67 
& 17.91 & 21.19 & 17.78 & 13.12 & 6.06 
& 19.00 & 24.30 & 20.46 & 15.09 & 7.50 \\

& 2 
& 44.95 & 44.08 & 39.05 & 33.15 & 19.86 
& 37.33 & 38.70 & 33.43 & 26.79 & 16.30 
& 13.54 & 19.15 & 15.14 & 9.63 & 3.99 
& 12.69 & \second{18.10} & 13.60 & 8.19 & 3.73 
& 4.80 & 11.48 & 5.87 & 3.86 & 1.90 
& 6.84 & 13.35 & 6.05 & 3.95 & 1.58 \\

& 3 
& 20.18 & 33.08 & 21.21 & 16.87 & 9.63 
& 22.03 & 32.15 & 21.50 & 17.57 & 9.81 
& 13.59 & 16.61 & 9.03 & 5.28 & 1.87 
& 13.37 & 15.78 & 8.86 & 5.02 & 1.97 
& 8.37 & 11.32 & 6.54 & 4.12 & 1.63 
& 7.60 & 13.07 & 6.66 & 4.23 & 1.68 \\

& 4 
& 28.68 & 32.10 & 21.78 & 16.37 & 8.39 
& \second{38.73} & \second{39.50} & \second{30.81} & \second{25.60} & \second{16.10} 
& 2.32 & 5.89 & 2.89 & 1.53 & 0.53 
& 4.69 & 8.85 & 5.17 & 3.31 & 1.43 
& 4.76 & 10.35 & 4.91 & 2.94 & 1.11 
& 5.73 & 11.09 & 6.12 & 3.82 & 1.28 \\

\midrule

\multirow{4}{*}{FACT~\cite{lu2024fact}}
& 1 
& \best{72.82} & \best{71.91} & \best{71.49} & \best{68.30} & \best{58.38} 
& \best{73.30} & \best{79.45} & \best{77.46} & \best{76.10} & \best{71.52} 
& \second{57.11} & \best{41.15} & \best{40.08} & \best{35.29} & \best{27.97} 
& \best{74.13} & \best{66.14} & \best{65.76} & \best{63.36} & \best{58.09} 
& \best{55.03} & \best{51.91} & \best{56.12} & \best{53.19} & \best{45.68} 
& \best{71.08} & \best{68.28} & \best{72.64} & \best{71.01} & \best{67.20} \\

& 2 
& \second{63.41} & \second{55.27} & \best{57.14} & \best{53.76} & \best{41.27} 
& \best{56.44} & \second{52.93} & \best{55.10} & \best{52.49} & \best{42.73} 
& 22.74 & \second{21.08} & \second{22.65} & \second{19.43} & \second{11.46} 
& 21.99 & 18.09 & \second{15.51} & \second{12.55} & \second{6.85} 
& \second{19.77} & \second{19.90} & \second{19.15} & \second{15.14} & \second{9.16} 
& \second{16.34} & \second{18.15} & \second{17.04} & \second{13.29} & \second{7.66} \\

& 3 
& 29.54 & \second{35.25} & 29.25 & 25.31 & 16.61 
& 27.15 & \second{34.32} & 26.59 & 23.00 & 15.86 
& \second{44.90} & \second{22.02} & \second{17.30} & \second{12.76} & \second{6.69} 
& 41.65 & \second{21.88} & 15.86 & 12.08 & 5.77 
& \second{19.00} & \second{15.29} & \second{13.45} & \second{10.46} & \second{6.35} 
& 21.95 & 15.42 & 12.84 & 9.94 & 5.40 \\

& 4 
& \second{37.19} & \second{35.31} & \second{25.48} & \second{20.45} & \second{11.82} 
& 36.81 & 37.30 & 28.04 & 23.93 & 14.86 
& 41.47 & 15.47 & 7.64 & 3.40 & 1.07 
& \second{48.92} & 17.21 & 8.10 & 4.05 & 1.27 
& \second{16.92} & 12.82 & 6.63 & 4.77 & 2.34 
& \second{21.28} & \second{15.50} & 9.88 & 6.49 & 2.91 \\

\bottomrule
\end{tabular}}
\end{table*}

\begin{table*}[t]
\centering
\caption{Cross-View Understanding Benchmarks: Temporal Correspondence (CV-TA) and Semantic Matching (CV-SM)}
\vskip-3mm
\label{tab:unified_cross_view_benchmark_full}
\setlength{\tabcolsep}{3.2pt}
\renewcommand{\arraystretch}{1.12}
\resizebox{0.95\textwidth}{!}{
\begin{tabular}{@{}l ccc ccc ccc ccc ccc ccc ccc cc cc cc cc@{}}
\toprule
\multirow{3}{*}{\textbf{Backbone}} &
\multicolumn{9}{c}{\textbf{CV-TA (cosine kNN)}} &
\multicolumn{12}{c}{\textbf{CV-SM Retrieval (cosine kNN)}} &
\multicolumn{8}{c}{\textbf{CV-SM Classification (linear probe)}} \\
\cmidrule(lr){2-10} \cmidrule(lr){11-22} \cmidrule(l){23-30}
&
\multicolumn{3}{c}{\textbf{Local}} &
\multicolumn{3}{c}{\textbf{Global}} &
\multicolumn{3}{c}{\textbf{Exo$\rightarrow$Ego}} &
\multicolumn{3}{c}{\textbf{Split 1}} &
\multicolumn{3}{c}{\textbf{Split 2}} &
\multicolumn{3}{c}{\textbf{Split 3}} &
\multicolumn{3}{c}{\textbf{Split 4}} &
\multicolumn{2}{c}{\textbf{Split 1}} &
\multicolumn{2}{c}{\textbf{Split 2}} &
\multicolumn{2}{c}{\textbf{Split 3}} &
\multicolumn{2}{c}{\textbf{Split 4}} \\
\cmidrule(lr){2-4}
\cmidrule(lr){5-7}
\cmidrule(lr){8-10}
\cmidrule(lr){11-13}
\cmidrule(lr){14-16}
\cmidrule(lr){17-19}
\cmidrule(lr){20-22}
\cmidrule(lr){23-24}
\cmidrule(lr){25-26}
\cmidrule(lr){27-28}
\cmidrule(l){29-30}
&
\textbf{R@1$\uparrow$} & \textbf{R@5$\uparrow$} & \textbf{MdR$\downarrow$} &
\textbf{R@1$\uparrow$} & \textbf{R@5$\uparrow$} & \textbf{MdR$\downarrow$} & 
\textbf{R@1$\uparrow$} & \textbf{R@5$\uparrow$} & \textbf{MdR$\downarrow$} & 
\textbf{R@1$\uparrow$} & \textbf{R@5$\uparrow$} & \textbf{mAP$\uparrow$} &
\textbf{R@1$\uparrow$} & \textbf{R@5$\uparrow$} & \textbf{mAP$\uparrow$} &
\textbf{R@1$\uparrow$} & \textbf{R@5$\uparrow$} & \textbf{mAP$\uparrow$} &
\textbf{R@1$\uparrow$} & \textbf{R@5$\uparrow$} & \textbf{mAP$\uparrow$} &
\textbf{Top-1$\uparrow$} & \textbf{MF1$\uparrow$} &
\textbf{Top-1$\uparrow$} & \textbf{MF1$\uparrow$} &
\textbf{Top-1$\uparrow$} & \textbf{MF1$\uparrow$} &
\textbf{Top-1$\uparrow$} & \textbf{MF1$\uparrow$} \\
\midrule
I3D~\cite{carreira2017quo}
& \second{4.95} & \second{21.50} & \second{13.67} 
& \second{0.69} & \second{2.24} & 333.00 
& \second{3.79} & \second{18.86} & \second{14.17} 
& \second{5.46} & \second{22.56} & \second{5.36}
& \second{5.08} & \second{20.90} & \second{5.53}
& 5.53 & 24.03 & 6.78
& 5.78 & \second{22.08} & \second{5.41}
& \second{39.99} & \second{40.41}
& \second{40.98} & \second{41.88}
& \second{18.01} & \second{14.57}
& \second{41.62} & \second{41.64} \\
VideoMAEv2~\cite{wang2023videomaev2}
& 4.02 & 19.36 & \second{13.67}
& 0.32 & 1.41 & \second{324.08}
& 3.43 & 18.65 & 14.33
& 5.44 & 20.29 & 5.30
& \second{5.08} & 20.57 & 5.48
& \best{8.64} & \best{28.51} & \second{7.06}
& \second{6.11} & 21.21 & 5.38
& 24.31 & 23.86
& 19.75 & 18.81
& 15.01 & 11.89
& 26.39 & 25.18 \\
MViTv2~\cite{li2022mvitv2}
& \best{7.38} & \best{29.18} & \best{10.17}
& \best{1.40} & \best{5.35} & \best{162.25}
& \best{5.12} & \best{21.38} & \best{13.00}
& \best{6.24} & \best{24.47} & \best{5.54}
& \best{5.82} & \best{23.52} & \best{5.80}
& \second{6.37} & \second{25.21} & \best{7.13}
& \best{6.38} & \best{24.87} & \best{5.58}
& \best{54.10} & \best{53.97}
& \best{53.44} & \best{54.17}
& \best{23.35} & \best{20.51}
& \best{54.83} & \best{54.87} \\
\bottomrule
\end{tabular}
}
\end{table*}

\subsection{Temporal Understanding Benchmark}
\label{subsec:exp_temporal}
Table~\ref{tab:tas_unified} reveals two consistent patterns. 
Moving from coarse step-level to bimanual atomic segmentation 
causes a steep performance drop: on S1, the best TAS-S result 
reaches F1@50$=$71.52 whereas the best TAS-BL result reaches only 27.97 (FACT+I3D) — a $2.6\times$ gap despite each hand operating over a smaller label space, suggesting that fine-grained hand-specific interactions are substantially harder to localize than coarse procedural steps. 
The two hands differ not only in performance but also in error mode: 
the left hand exhibits more stable temporal behavior (Acc.), 
while the right hand exhibits stronger boundary-localized detection 
(F1), indicating a structural asymmetry between stabilizing and 
tool-driving roles that recurs in Sec.~\ref{subsec:exp_forecasting}. 
Across splits, ASQuery remains relatively stable under viewpoint and 
configuration shift (I3D, S1$\to$S4: 33.89$\to$38.56) while 
FACT drops sharply under cross-configuration (S3: 15.86), 
showing that strong IID performance does not guarantee 
deployment-oriented generalization.

\subsection{Cross-view Understanding Benchmark}
\label{subsec:exp_crossview}
Table~\ref{tab:unified_cross_view_benchmark_full} reveals 
a structural limitation between what exocentric and 
egocentric streams can jointly resolve, arising from 
two separable sources. Instance alignment and semantic matching 
diverge sharply: MViTv2 achieves CV-SM retrieval R@1$=$6.24 
yet global CV-TA R@1$=$1.40 using identical frozen 
representations — a $4\times$ gap arising because same-instance 
segments across views are more visually dissimilar than 
same-class segments across trials, suggesting that ego-exo 
representation learning requires instance-level temporal 
objectives beyond semantic correspondence. The 
Exo$\rightarrow$Ego setting then quantifies the hard ceiling: 
MViTv2 Local R@1 falls from 7.38 to 5.12, and Global R@1 
collapses to 1.40, suggesting that the gap is not solely a 
representation-learning issue but is also bounded by 
view-dependent missing evidence caused by hand and tool 
occlusion. A secondary finding concerns pretraining objective: 
MViTv2 leads classification by $2\times$ over VideoMAEv2 
(54.10 vs.\ 24.31, S1) while retrieval gaps are far smaller 
(6.24 vs.\ 5.44), consistent with masked reconstruction 
yielding representations suited for appearance similarity 
but less calibrated to the semantic boundaries that separate 
assembly action classes.

\subsection{Action Forecasting Benchmark}
\label{subsec:exp_forecasting}
Table~\ref{tab:af_main} reveals that the failure modes
established in segmentation persist and sharpen under
anticipation. Bimanual asymmetry persists in short-term anticipation 
(ScalAnt+I3D: left 18.36 vs.\ right 15.70), 
indicating systematically different predictability: the stabilizing 
hand is more temporally persistent and thus easier to forecast, 
while the tool-driving hand involves sharper, decision-dependent 
transitions. This does not imply full observability, as the supporting 
role remains weakly constrained by explicit visual cues. A noun--action gap (AVT+I3D: noun 54.20\% vs.\ act 17.34\%) further isolates
the bottleneck: object identity is appearance-predictable,
manipulation intent is not. 
At the long horizon, ScalAnt establishes a consistent visual ceiling 
(I3D: AUED$=$0.622; VideoMAEv2: 0.644) that neither language-model 
augmentation nor zero-shot reasoning can breach: AntGPT's fine-tuned 
Llama2~\cite{touvron2023llama} reaches only 0.667 despite operating 
on recognized step sequences rather than raw video, and PALM's few-shot 
In-Context Learning degrades further to 0.801, suggesting that limited 
upstream step recognition constrains the effectiveness of symbolic 
priors in end-to-end deployment. Beyond 2--3 steps, topological ambiguity 
from IMPACT's multi-route prerequisite graph dominates over both visual 
and linguistic signal, identifying graph-aware reasoning as the key 
missing ingredient rather than model capacity. Qwen3VL-8B collapses 
across both horizons (AF-S act: 4.86\%; AF-L AUED: 0.827) while retaining 
noun recognition (32.14\%), a dissociation revealing that recognizing 
what objects are present does not transfer to reasoning about how they 
should be manipulated.

\begin{table}[t]
\centering
\caption{Action Forecasting Benchmarks.}
\label{tab:af_main}
\vskip -3mm
\scriptsize
\setlength{\tabcolsep}{3.4pt}
\renewcommand{\arraystretch}{1.10}

\textbf{(a) Short-term Anticipation (AF-S)}
\vskip 1mm

\resizebox{0.9\columnwidth}{!}{
\begin{tabular}{llccccccccc}
\toprule
\multirow{2}{*}{\textbf{Model}} & \multirow{2}{*}{\makecell{\textbf{Feat.}\\\textbf{Ext.}}}
& \multicolumn{3}{c}{\textbf{Overall}}
& \multicolumn{3}{c}{\textbf{Left}}
& \multicolumn{3}{c}{\textbf{Right}} \\
\cmidrule(lr){3-5} \cmidrule(lr){6-8} \cmidrule(lr){9-11}
& & \textbf{act}$\uparrow$ & \textbf{verb}$\uparrow$ & \textbf{noun}$\uparrow$
  & \textbf{act}$\uparrow$ & \textbf{verb}$\uparrow$ & \textbf{noun}$\uparrow$
  & \textbf{act}$\uparrow$ & \textbf{verb}$\uparrow$ & \textbf{noun}$\uparrow$ \\
\midrule
\multirow{2}{*}{AVT~\cite{girdhar2021anticipative}}
& I3D~\cite{carreira2017quo}
& \second{17.34} & \second{42.61} & \best{54.20}
& \second{17.33} & \second{45.32} & \best{50.81}
& \second{15.10} & \second{38.28} & \best{50.95} \\
\cmidrule(lr){2-11}
& VideoMAEv2~\cite{wang2023videomaev2}
& 10.61 & 33.10 & 42.95
& 10.70 & 32.90 & 39.77
& 10.40 & 30.97 & 40.49 \\
\midrule
\multirow{2}{*}{ScalAnt~\cite{zhong2026scalable}}
& I3D~\cite{carreira2017quo}
& \best{18.41} & \best{44.44} & \second{53.42}
& \best{18.36} & \best{45.80} & \second{49.30}
& \best{15.70} & \best{40.92} & \second{50.52} \\
\cmidrule(lr){2-11}
& VideoMAEv2~\cite{wang2023videomaev2}
& 13.89 & 38.80 & 48.39
& 13.32 & 40.58 & 46.75
& 13.16 & 36.18 & 45.61 \\
\midrule
Qwen3VL-8B~\cite{bai2025qwen3}
& --
& 4.86 & 6.99 & 32.14
& 5.95 & 6.90 & 33.28
& 4.26 & 6.96 & 31.53 \\
\bottomrule
\end{tabular}
}

\vskip 2mm

\textbf{(b) Long-horizon Forecasting (AF-L)}
\vskip 1mm

\resizebox{0.9\columnwidth}{!}{
\begin{tabular}{llccccccc}
\toprule
\multirow{2}{*}{\textbf{Model}} & \multirow{2}{*}{\makecell{\textbf{Feat.}\\\textbf{Ext.}}}
& \multirow{2}{*}{\textbf{AUED}$\downarrow$}
& \multirow{2}{*}{\textbf{Acc@1}$\uparrow$}
& \multicolumn{5}{c}{\textbf{Per-Step ED}$\downarrow$} \\
\cmidrule(lr){5-9}
& & & & \textbf{@1} & \textbf{@2} & \textbf{@3} & \textbf{@4} & \textbf{@5} \\
\midrule
\multirow{2}{*}{ScalAnt~\cite{zhong2026scalable}}
& I3D~\cite{carreira2017quo}
& \best{0.622} & \second{26.92} & \best{0.554} & \best{0.595} & \best{0.631} & \best{0.654} & \best{0.666} \\
\cmidrule(lr){2-9}
& VideoMAEv2~\cite{wang2023videomaev2}
& \second{0.644} & \best{28.49} & \second{0.564} & \second{0.625} & \second{0.654} & \second{0.672} & \second{0.683} \\
\midrule
AntGPT~\cite{qi2024antgpt}
& I3D~\cite{carreira2017quo}
& 0.667 & 24.90 & 0.699 & 0.671 & 0.668 & 0.653 & 0.654 \\
\cmidrule(lr){2-9}
& VideoMAEv2~\cite{wang2023videomaev2}
& 0.693 & 23.33 & 0.737 & 0.686 & 0.696 & 0.680 & 0.683 \\
\midrule
PALM~\cite{kim2024palm}
& I3D~\cite{carreira2017quo}
& 0.801 & 17.89 & 0.817 & 0.831 & 0.797 & 0.777 & 0.781 \\
\cmidrule(lr){2-9}
& VideoMAEv2~\cite{wang2023videomaev2}
& 0.790 & 20.50 & 0.795 & 0.816 & 0.787 & 0.773 & 0.775 \\
\midrule
Qwen3VL-8B~\cite{bai2025qwen3}
& --
& 0.827 & 15.89 & 0.841 & 0.834 & 0.826 & 0.818 & 0.818 \\
\bottomrule
\end{tabular}
}
\vskip -2.0em
\end{table}

\subsection{State \& Reasoning Benchmark}
\label{subsec:exp_reasoning}
Table~\ref{tab:proc_main} concentrates all three preceding 
gaps at the operationally critical moments of real assembly. 
MS-TCN++ achieves Final-Acc of 0.93 yet Trans-F1 of 0.33 — 
the best across all methods — showing that discriminative 
models learn stable component-state distributions but miss 
the brief transitions where monitoring matters most, 
mirroring the generalization failure of 
Sec.~\ref{subsec:exp_temporal}. On PSR, Gemini 3.1 Pro 
leads on procedural ordering (POS: 0.36 vs.\ 0.21) but 
incurs more than twice the detection delay (19.26\,s 
vs.\ 8.46\,s), instantiating what we term the \textit{knowledge--execution 
gap}: vision-language models carry strong procedural 
ordering priors yet cannot ground completion events 
in time. Even when overall PPR accuracy is high for 
stronger baselines, recovery remains almost unresolved: 
F1$_\text{rec}$ stays at or near zero on the left hand 
across all models and below 7 on the right, while DiffAct's 
PPR-L anomaly F1 (13.60\%) at near-zero accuracy (6.03\%) 
strongly suggests that rare-phase imbalance — recovery 
spans only 2.22\% of frames — is a major bottleneck rather 
than model capacity alone. The observability gap, the 
graph-structural forecasting ceiling, and the knowledge--execution gap each localize to the same operationally critical 
moments: transitions, anomalies, and recovery.

\begin{table}[t]
\centering
\caption{\textbf{State \& Reasoning Benchmarks.}}
\label{tab:proc_main}
\vskip -3mm
\scriptsize
\setlength{\tabcolsep}{3.4pt}
\renewcommand{\arraystretch}{1.10}

\textbf{(a) Procedure Step Recognition (PSR) \& Assembly State Recognition (ASR)}
\vskip 1mm

\resizebox{0.9\columnwidth}{!}{
\begin{tabular}{lcccccc}
\toprule
\multirow{2}{*}{\textbf{Method}} &
\multicolumn{3}{c}{\textbf{ASR Metrics}} &
\multicolumn{3}{c}{\textbf{PSR Metrics}} \\
\cmidrule(lr){2-4}\cmidrule(lr){5-7}
& \textbf{Macro-F1}$\uparrow$ & \textbf{Trans-F1}$\uparrow$ & \textbf{Final-Acc}$\uparrow$
& \textbf{POS}$\uparrow$ & \textbf{F1}$\uparrow$ & \textbf{$\tau$ (s)}$\downarrow$ \\
\midrule
MS-TCN++~\cite{li2020ms}
& \best{0.84} & \best{0.33} & \best{0.93}
& \second{0.21} & \best{0.26} & \best{8.46} \\
VideoMAEv2~\cite{wang2023videomaev2}
& 0.38 & 0.01 & 0.33
& 0.00 & 0.01 & 24.20 \\
STORM-PSR~\cite{schoonbeek2025learning}
& \second{0.46} & 0.03 & 0.51
& 0.00 & 0.01 & 22.66 \\
Gemini 3.1 Pro~\cite{team2023gemini}
& 0.44 & \second{0.10} & \second{0.81}
& \best{0.36} & \second{0.23} & \second{19.26} \\
\bottomrule
\end{tabular}
}

\vskip 2mm

\textbf{(b) Procedural Phase Recognition (PPR) \& Anomaly Type Recognition (ATR)}
\vskip 1mm

\resizebox{0.9\columnwidth}{!}{
\begin{tabular}{lcccccccccc}
\toprule
\multirow{2}{*}{\textbf{Model}} &
\multicolumn{4}{c}{\textbf{PPR-L}} &
\multicolumn{4}{c}{\textbf{PPR-R}} &
\textbf{ATR-L} & \textbf{ATR-R} \\
\cmidrule(lr){2-5}\cmidrule(lr){6-9}
& \textbf{Acc.}$\uparrow$ & \textbf{Macro-F1}$\uparrow$ & \textbf{F1$_{\text{anom}}$}$\uparrow$ & \textbf{F1$_{\text{rec}}$}$\uparrow$
& \textbf{Acc.}$\uparrow$ & \textbf{Macro-F1}$\uparrow$ & \textbf{F1$_{\text{anom}}$}$\uparrow$ & \textbf{F1$_{\text{rec}}$}$\uparrow$
& \textbf{mAP}$\uparrow$ & \textbf{mAP}$\uparrow$ \\
\midrule
ASQuery~\cite{gan2024asquery}
& 84.14 & 31.22 & 2.30 & \second{0.00}
& 74.50 & \second{33.79} & 15.08 & \second{1.22}
& -- & -- \\
DiffAct~\cite{liu2023diffusion}
& 6.03 & 5.16 & \best{13.60} & \best{1.25}
& 48.18 & 27.96 & \best{20.97} & \best{6.41}
& -- & -- \\
LTContext~\cite{bahrami2023much}
& \best{90.10} & \second{31.46} & 0.00 & \second{0.00}
& \best{87.41} & 30.88\ & 0.00 & 0.00
& \second{27.43} & \second{29.22} \\
FACT~\cite{lu2024fact}
& \second{85.70} & \best{32.62} & \second{5.59} & \second{0.00}
& \second{81.70} & \best{35.03} & \second{16.30} & 0.00
& \best{31.86} & \best{31.93} \\
\bottomrule
\end{tabular}
}

\vskip -1.0em
\end{table}

\section{Conclusion}
IMPACT is the first dataset to simultaneously provide 
synchronized ego--exo capture, bimanual annotation, 
compliance-aware state tracking, and anomaly--recovery 
supervision within a real commercial power-tool workflow.
Evaluation across four task families surfaces three 
challenges that converge on the same operationally 
critical moments of transitions, anomalies, and recovery: 
the egocentric observability gap that view-invariant 
pretraining cannot resolve; the graph-structural forecasting 
ceiling imposed by execution-path ambiguity rather than 
model capacity; and the knowledge--execution gap through 
which vision-language models fail to translate object 
recognition into manipulation reasoning.
That all three localize to the same moments is itself 
a finding, with direct implications for temporal action 
segmentation, cross-view representation learning, 
procedural video understanding, anomaly detection, 
and human-robot collaboration. Deployment-grade assembly understanding requires not stronger models on individual tasks, but joint reasoning 
over actions, states, and corrective behavior under 
conditions that only a unified benchmark can impose.

\subsection*{Ethics, Access, and Reproducibility}
All participants provided informed consent; no personally 
identifiable information is included in the released data, 
and all recordings were anonymized following standard 
research ethics practices. IMPACT is released under 
CC BY 4.0, permitting free academic and research use. 
The full dataset, annotations, evaluation code, and 
baseline implementations are available at 
\url{https://github.com/Kratos-Wen/IMPACT}; detailed 
hyperparameters and training commands are provided in 
the supplementary to support exact replication.

\begin{acks}
This work was supported in part by the Deutsche Forschungsgemeinschaft (DFG, German Research Foundation) - SFB 1574 - 471687386. 
\end{acks}

\clearpage
\bibliographystyle{ACM-Reference-Format}
\bibliography{acmart}

@article{wen2025mica,
  title={Mica: Multi-agent industrial coordination assistant},
  author={Wen, Di and Peng, Kunyu and Zheng, Junwei and Chen, Yufan and Shi, Yitian and Wei, Jiale and Liu, Ruiping and Yang, Kailun and Stiefelhagen, Rainer},
  journal={arXiv preprint arXiv:2509.15237},
  year={2025}
}

@article{wen2025snap,
  title={Snap, segment, deploy: A visual data and detection pipeline for wearable industrial assistants},
  author={Wen, Di and Zheng, Junwei and Liu, Ruiping and Xu, Yi and Peng, Kunyu and Stiefelhagen, Rainer},
  journal={arXiv preprint arXiv:2507.21072},
  year={2025}
}

@article{ragusa2023meccano,
  title={Meccano: A multimodal egocentric dataset for humans behavior understanding in the industrial-like domain},
  author={Ragusa, Francesco and Furnari, Antonino and Farinella, Giovanni Maria},
  journal={Computer vision and image understanding},
  volume={235},
  pages={103764},
  year={2023},
  publisher={Elsevier}
}

@inproceedings{ragusa2024enigma,
  title={Enigma-51: Towards a fine-grained understanding of human behavior in industrial scenarios},
  author={Ragusa, Francesco and Leonardi, Rosario and Mazzamuto, Michele and Bonanno, Claudia and Scavo, Rosario and Furnari, Antonino and Farinella, Giovanni Maria},
  booktitle={Proceedings of the IEEE/CVF Winter Conference on Applications of Computer Vision},
  pages={4549--4559},
  year={2024}
}

@inproceedings{wang2023holoassist,
  title={Holoassist: an egocentric human interaction dataset for interactive ai assistants in the real world},
  author={Wang, Xin and Kwon, Taein and Rad, Mahdi and Pan, Bowen and Chakraborty, Ishani and Andrist, Sean and Bohus, Dan and Feniello, Ashley and Tekin, Bugra and Frujeri, Felipe Vieira and others},
  booktitle={Proceedings of the IEEE/CVF International Conference on Computer Vision},
  pages={20270--20281},
  year={2023}
}

@inproceedings{li2021ego,
  title={Ego-exo: Transferring visual representations from third-person to first-person videos},
  author={Li, Yanghao and Nagarajan, Tushar and Xiong, Bo and Grauman, Kristen},
  booktitle={Proceedings of the IEEE/CVF Conference on Computer Vision and Pattern Recognition},
  pages={6943--6953},
  year={2021}
}

@article{aganian2023attach,
  title={Attach dataset: Annotated two-handed assembly actions for human action understanding},
  author={Aganian, Dustin and Stephan, Benedict and Eisenbach, Markus and Stretz, Corinna and Gross, Horst-Michael},
  journal={arXiv preprint arXiv:2304.08210},
  year={2023}
}

@inproceedings{schoonbeek2024industreal,
  title={Industreal: A dataset for procedure step recognition handling execution errors in egocentric videos in an industrial-like setting},
  author={Schoonbeek, Tim J and Houben, Tim and Onvlee, Hans and Van der Sommen, Fons and others},
  booktitle={Proceedings of the IEEE/CVF Winter Conference on Applications of Computer Vision},
  pages={4365--4374},
  year={2024}
}

@article{chavan2025indego,
  title={IndEgo: A Dataset of Industrial Scenarios and Collaborative Work for Egocentric Assistants},
  author={Chavan, Vivek and Imgrund, Yasmina and Dao, Tung and Bai, Sanwantri and Wang, Bosong and Lu, Ze and Heimann, Oliver and Kr{\"u}ger, J{\"o}rg},
  journal={arXiv preprint arXiv:2511.19684},
  year={2025}
}

@article{cicirelli2022ha4m,
  title={The HA4M dataset: Multi-Modal Monitoring of an assembly task for Human Action recognition in Manufacturing},
  author={Cicirelli, Grazia and Marani, Roberto and Romeo, Laura and Dom{\'\i}nguez, Manuel Garc{\'\i}a and Heras, J{\'o}nathan and Perri, Anna G and D’Orazio, Tiziana},
  journal={Scientific Data},
  volume={9},
  number={1},
  pages={745},
  year={2022},
  publisher={Nature Publishing Group UK London}
}

@inproceedings{flaborea2024prego,
  title={Prego: online mistake detection in procedural egocentric videos},
  author={Flaborea, Alessandro and Di Melendugno, Guido Maria D'Amely and Plini, Leonardo and Scofano, Luca and De Matteis, Edoardo and Furnari, Antonino and Farinella, Giovanni Maria and Galasso, Fabio},
  booktitle={Proceedings of the IEEE/CVF Conference on Computer Vision and Pattern Recognition},
  pages={18483--18492},
  year={2024}
}

@inproceedings{quattrocchi2024synchronization,
  title={Synchronization is all you need: Exocentric-to-egocentric transfer for temporal action segmentation with unlabeled synchronized video pairs},
  author={Quattrocchi, Camillo and Furnari, Antonino and Di Mauro, Daniele and Giuffrida, Mario Valerio and Farinella, Giovanni Maria},
  booktitle={European Conference on Computer Vision},
  pages={253--270},
  year={2024},
  organization={Springer}
}

@inproceedings{huang2024egoexolearn,
  title={Egoexolearn: A dataset for bridging asynchronous ego-and exo-centric view of procedural activities in real world},
  author={Huang, Yifei and Chen, Guo and Xu, Jilan and Zhang, Mingfang and Yang, Lijin and Pei, Baoqi and Zhang, Hongjie and Dong, Lu and Wang, Yali and Wang, Limin and others},
  booktitle={Proceedings of the IEEE/CVF Conference on Computer Vision and Pattern Recognition},
  pages={22072--22086},
  year={2024}
}

@inproceedings{lee2024error,
  title={Error detection in egocentric procedural task videos},
  author={Lee, Shih-Po and Lu, Zijia and Zhang, Zekun and Hoai, Minh and Elhamifar, Ehsan},
  booktitle={Proceedings of the IEEE/CVF Conference on Computer Vision and Pattern Recognition},
  pages={18655--18666},
  year={2024}
}

@article{peddi2024captaincook4d,
  title={Captaincook4d: A dataset for understanding errors in procedural activities},
  author={Peddi, Rohith and Arya, Shivvrat and Challa, Bharath and Pallapothula, Likhitha and Vyas, Akshay and Gouripeddi, Bhavya and Zhang, Qifan and Wang, Jikai and Komaragiri, Vasundhara and Ragan, Eric and others},
  journal={Advances in Neural Information Processing Systems},
  volume={37},
  pages={135626--135679},
  year={2024}
}

@inproceedings{lee2025error,
  title={Error recognition in procedural videos using generalized task graph},
  author={Lee, Shih-Po and Elhamifar, Ehsan},
  booktitle={Proceedings of the IEEE/CVF International Conference on Computer Vision},
  pages={10009--10021},
  year={2025}
}

@inproceedings{huang2025modeling,
  title={Modeling multiple normal action representations for error detection in procedural tasks},
  author={Huang, Wei-Jin and Li, Yuan-Ming and Xia, Zhi-Wei and Tang, Yu-Ming and Lin, Kun-Yu and Hu, Jian-Fang and Zheng, Wei-Shi},
  booktitle={Proceedings of the Computer Vision and Pattern Recognition Conference},
  pages={27794--27804},
  year={2025}
}

@inproceedings{xue2024learning,
  title={Learning object state changes in videos: An open-world perspective},
  author={Xue, Zihui and Ashutosh, Kumar and Grauman, Kristen},
  booktitle={Proceedings of the IEEE/CVF Conference on Computer Vision and Pattern Recognition},
  pages={18493--18503},
  year={2024}
}

@inproceedings{lea2017temporal,
  title={Temporal convolutional networks for action segmentation and detection},
  author={Lea, Colin and Flynn, Michael D and Vidal, Rene and Reiter, Austin and Hager, Gregory D},
  booktitle={proceedings of the IEEE Conference on Computer Vision and Pattern Recognition},
  pages={156--165},
  year={2017}
}

@inproceedings{farha2019ms,
  title={Ms-tcn: Multi-stage temporal convolutional network for action segmentation},
  author={Farha, Yazan Abu and Gall, Jurgen},
  booktitle={Proceedings of the IEEE/CVF conference on computer vision and pattern recognition},
  pages={3575--3584},
  year={2019}
}

@article{li2020ms,
   author={Shi-Jie Li and Yazan AbuFarha and Yun Liu and Ming-Ming Cheng and Juergen Gall},
    journal={IEEE Transactions on Pattern Analysis and Machine Intelligence}, 
    title={MS-TCN++: Multi-Stage Temporal Convolutional Network for Action Segmentation}, 
    year={2020},
    volume={},
    number={},
    pages={1-1},
    doi={10.1109/TPAMI.2020.3021756},
}

@InProceedings{wang2023videomaev2,
    author    = {Wang, Limin and Huang, Bingkun and Zhao, Zhiyu and Tong, Zhan and He, Yinan and Wang, Yi and Wang, Yali and Qiao, Yu},
    title     = {VideoMAE V2: Scaling Video Masked Autoencoders With Dual Masking},
    booktitle = {Proceedings of the IEEE/CVF Conference on Computer Vision and Pattern Recognition (CVPR)},
    month     = {June},
    year      = {2023},
    pages     = {14549-14560}
}

@article{damen2022rescaling,
   title={Rescaling Egocentric Vision},
   author={Damen, Dima and Doughty, Hazel and Farinella, Giovanni Maria  and and Furnari, Antonino 
           and Ma, Jian and Kazakos, Evangelos and Moltisanti, Davide and Munro, Jonathan 
           and Perrett, Toby and Price, Will and Wray, Michael},
           journal={International Journal of Computer Vision},
           volume={130},
           number={1},
           pages={33--55},
           year={2022},
           publisher={Springer}
}

@inproceedings{grauman2022ego4d,
  title={Ego4d: Around the world in 3,000 hours of egocentric video},
  author={Grauman, Kristen and Westbury, Andrew and Byrne, Eugene and Chavis, Zachary and Furnari, Antonino and Girdhar, Rohit and Hamburger, Jackson and Jiang, Hao and Liu, Miao and Liu, Xingyu and others},
  booktitle={Proceedings of the IEEE/CVF conference on computer vision and pattern recognition},
  pages={18995--19012},
  year={2022}
}

@inproceedings{sener2022assembly101,
  title={Assembly101: A large-scale multi-view video dataset for understanding procedural activities},
  author={Sener, Fadime and Chatterjee, Dibyadip and Shelepov, Daniel and He, Kun and Singhania, Dipika and Wang, Robert and Yao, Angela},
  booktitle={CVPR},
  year={2022}
}

@article{zheng2023ha,
  title={Ha-vid: A human assembly video dataset for comprehensive assembly knowledge understanding},
  author={Zheng, Hao and Lee, Regina and Lu, Yuqian},
  journal={NeurIPS},
  year={2023}
}

@inproceedings{ben2021ikea,
  title={The ikea asm dataset: Understanding people assembling furniture through actions, objects and pose},
  author={Ben-Shabat, Yizhak and Yu, Xin and Saleh, Fatemeh and Campbell, Dylan and Rodriguez-Opazo, Cristian and Li, Hongdong and Gould, Stephen},
  booktitle={WACV},
  year={2021}
}

@inproceedings{grauman2024ego,
  title={Ego-exo4d: Understanding skilled human activity from first-and third-person perspectives},
  author={Grauman, Kristen and Westbury, Andrew and Torresani, Lorenzo and Kitani, Kris and Malik, Jitendra and Afouras, Triantafyllos and Ashutosh, Kumar and Baiyya, Vijay and Bansal, Siddhant and Boote, Bikram and others},
  booktitle={CVPR},
  year={2024}
}

@inproceedings{bahrami2023much,
  title={How much temporal long-term context is needed for action segmentation?},
  author={Bahrami, Emad and Francesca, Gianpiero and Gall, Juergen},
  booktitle={Proceedings of the IEEE/CVF International Conference on Computer Vision},
  pages={10351--10361},
  year={2023}
}

@inproceedings{gan2024asquery,
  title={ASQuery: A query-based model for action segmentation},
  author={Gan, Ziliang and Jin, Lei and Nie, Lei and Wang, Zheng and Zhou, Li and Li, Liang and Wang, Zhecan and Li, Jianshu and Xing, Junliang and Zhao, Jian},
  booktitle={2024 IEEE International Conference on Multimedia and Expo (ICME)},
  pages={i--vi},
  year={2024},
  organization={IEEE}
}

@inproceedings{liu2023diffusion,
  title={Diffusion action segmentation},
  author={Liu, Daochang and Li, Qiyue and Dinh, Anh-Dung and Jiang, Tingting and Shah, Mubarak and Xu, Chang},
  booktitle={Proceedings of the IEEE/CVF international conference on computer vision},
  pages={10139--10149},
  year={2023}
}

@inproceedings{lu2024fact,
  title={Fact: Frame-action cross-attention temporal modeling for efficient action segmentation},
  author={Lu, Zijia and Elhamifar, Ehsan},
  booktitle={Proceedings of the IEEE/CVF Conference on Computer Vision and Pattern Recognition},
  pages={18175--18185},
  year={2024}
}

@inproceedings{li2022mvitv2,
  title={Mvitv2: Improved multiscale vision transformers for classification and detection},
  author={Li, Yanghao and Wu, Chao-Yuan and Fan, Haoqi and Mangalam, Karttikeya and Xiong, Bo and Malik, Jitendra and Feichtenhofer, Christoph},
  booktitle={Proceedings of the IEEE/CVF conference on computer vision and pattern recognition},
  pages={4804--4814},
  year={2022}
}

@inproceedings{carreira2017quo,
  title={Quo vadis, action recognition? a new model and the kinetics dataset},
  author={Carreira, Joao and Zisserman, Andrew},
  booktitle={proceedings of the IEEE Conference on Computer Vision and Pattern Recognition},
  pages={6299--6308},
  year={2017}
}

@inproceedings{park2025bootstrap,
  title={Bootstrap your own views: Masked ego-exo modeling for fine-grained view-invariant video representations},
  author={Park, Jungin and Lee, Jiyoung and Sohn, Kwanghoon},
  booktitle={Proceedings of the IEEE/CVF Conference on Computer Vision and Pattern Recognition},
  pages={13661--13670},
  year={2025}
}

@article{xue2023learning,
  title={Learning fine-grained view-invariant representations from unpaired ego-exo videos via temporal alignment},
  author={Xue, Zihui Sherry and Grauman, Kristen},
  journal={Advances in Neural Information Processing Systems},
  volume={36},
  pages={53688--53710},
  year={2023}
}

@inproceedings{girdhar2021anticipative,
  title={Anticipative video transformer},
  author={Girdhar, Rohit and Grauman, Kristen},
  booktitle={Proceedings of the IEEE/CVF international conference on computer vision},
  pages={13505--13515},
  year={2021}
}

@inproceedings{zhong2026scalable,
  title={Scalable Video Action Anticipation with Cross Linear Attentive Memory},
  author={Zhong, Zeyun and Martin, Manuel and Schneider, David and Lerch, David J and Wu, Chengzhi and Diederichs, Frederik and Gall, Juergen and Beyerer, J{\"u}rgen},
  booktitle={Proceedings of the IEEE/CVF Winter Conference on Applications of Computer Vision},
  pages={8113--8123},
  year={2026}
}

@article{bai2025qwen3,
  title={Qwen3-vl technical report},
  author={Bai, Shuai and Cai, Yuxuan and Chen, Ruizhe and Chen, Keqin and Chen, Xionghui and Cheng, Zesen and Deng, Lianghao and Ding, Wei and Gao, Chang and Ge, Chunjiang and others},
  journal={arXiv preprint arXiv:2511.21631},
  year={2025}
}

@inproceedings{damen2018scaling,
  title={Scaling egocentric vision: The epic-kitchens dataset},
  author={Damen, Dima and Doughty, Hazel and Farinella, Giovanni Maria and Fidler, Sanja and Furnari, Antonino and Kazakos, Evangelos and Moltisanti, Davide and Munro, Jonathan and Perrett, Toby and Price, Will and others},
  booktitle={Proceedings of the European conference on computer vision (ECCV)},
  pages={720--736},
  year={2018}
}

@inproceedings{ben2024ikea,
  title={Ikea ego 3d dataset: Understanding furniture assembly actions from ego-view 3d point clouds},
  author={Ben-Shabat, Yizhak and Paul, Jonathan and Segev, Eviatar and Shrout, Oren and Gould, Stephen},
  booktitle={Proceedings of the IEEE/CVF Winter Conference on Applications of Computer Vision},
  pages={4355--4364},
  year={2024}
}

@article{schoonbeek2025learning,
  title={Learning to recognize correctly completed procedure steps in egocentric assembly videos through spatio-temporal modeling},
  author={Schoonbeek, Tim J and Hung, Shao-Hsuan and Lehman, Dan and Onvlee, Hans and Kustra, Jacek and de With, Peter HN and Van der Sommen, Fons},
  journal={Computer Vision and Image Understanding},
  pages={104528},
  year={2025},
  publisher={Elsevier}
}

@article{team2023gemini,
  title={Gemini: a family of highly capable multimodal models},
  author={Team, Gemini and Anil, Rohan and Borgeaud, Sebastian and Alayrac, Jean-Baptiste and Yu, Jiahui and Soricut, Radu and Schalkwyk, Johan and Dai, Andrew M and Hauth, Anja and Millican, Katie and others},
  journal={arXiv preprint arXiv:2312.11805},
  year={2023}
}

@inproceedings{qi2024antgpt,
  title={{AntGPT}: Can Large Language Models Help Long-term Action Anticipation from Videos?},
  author={Qi Zhao and Shijie Wang and Ce Zhang and Changcheng Fu and Minh Quan Do and Nakul Agarwal and Kwonjoon Lee and Chen Sun},
  year={2024},
  booktitle={ICLR},
}

@inproceedings{kim2024palm,
  title={{PALM}: Predicting Actions through Language Models},
  author={Sanghwan Kim and Daoji Huang and Yongqin Xian and Otmar Hilliges and Luc Van Gool and Xi Wang},
  year={2024},
  booktitle={ECCV},
  pages={140--158},
  doi={10.1007/978-3-031-73007-8_9},
}

@article{touvron2023llama,
  title={{Llama 2}: Open Foundation and Fine-Tuned Chat Models},
  author={Hugo Touvron and Louis Martin and Kevin Stone and Peter Albert and others},
  journal={arXiv preprint arXiv:2307.09288},
  year={2023},
}


\end{document}